\documentclass[journal]{IEEEtran}
\usepackage{tikz}
\usepackage{amssymb}
\usepackage{booktabs}
\usepackage{amsmath}
\usepackage[linesnumbered,ruled]{algorithm2e}
\usepackage{algpseudocode}
\usepackage{caption}
\usepackage{tabulary}
\usepackage{array,booktabs,tabularx}

\usepackage{cite}

\usepackage{graphicx}
\usepackage{amssymb}
\usepackage{floatrow}
\usepackage{amsmath}
\usepackage{lipsum}
\usepackage{multicol}
\usepackage{tikz}
\newcommand{\nonl}{\renewcommand{\nl}{\let\nl\oldnl}}

\usetikzlibrary{arrows,shapes,positioning,shadows,trees}

\tikzset{
  basic/.style  = {draw, text width=2cm, drop shadow, font=\scriptsize, rectangle},
  root/.style   = {basic, rounded corners=2pt, thin, align=center,
                   fill=white!30},
  level 2/.style = {basic, rounded corners=4pt, thin,align=center, fill=white!60,
                   text width=5em},
  level 3/.style = {basic, thin, align=left, fill=white!60, text width=5em}
}

\newcommand{\CASE}[1]{\STATE \textbf{case} #1\textbf{:} \begin{ALC@g}}
\newcommand{\ENDCASE}{\end{ALC@g}}

\newcommand{\DEFAULT}{\STATE \textbf{default:} \begin{ALC@g}}
\newcommand{\ENDDEFAULT}{\end{ALC@g}}
\newcommand{\DEFAULTLINE}[1]{\STATE \textbf{default:} }

\ifCLASSINFOpdf
\else
\fi
\begin{document}
%
\title{ AI-based Modeling and Data-driven Evaluation for Smart Manufacturing Processes }
%
%

\author{Mohammadhossein Ghahramani,~\IEEEmembership{Member,~IEEE,}
           Yan~Qiao,~\IEEEmembership{Member,~IEEE,}\\
           MengChu~Zhou,~\IEEEmembership{Fellow,~IEEE},~
           Adrian~O'Hagan,
           and~James~Sweeney
\thanks{This work was supported by in part by Science and Technology development fund (FDCT) of Macau under Grant 011/2017/A, and in part by the National Natural Science Foundation of China under Grant 61803397.}
\thanks{}
\thanks{M. Ghahramani is with University College Dublin, Ireland, (e-mail: sphr.ghahramani@gmail.com, sepehr.ghahramani@ucd.ie).}
\thanks{Y. Qiao is with the Institute of Systems Engineering, Macau University of Science and Technology, Macau, (e-mail: yqiao@must.edu.mo).}
\thanks{M. C. Zhou is with the Helen and John C. Hartmann Department of Electrical and Computer Engineering, New Jersey Institute of Technology, Newark, NJ 07102, USA (e-mail: zhou@njit.edu).}
\thanks{A. O'Hagan is with University College Dublin, Ireland, (e-mail: adrian.ohagan@ucd.ie).}
\thanks{J. Sweeney is with the Royal College of Surgeons in Ireland (RCSI), Dublin, Ireland, (e-mail: james.sweeney@ucd.ie).}

}

%
%

\markboth{}%
{Shell \MakeLowercase{\textit{et al.}}: Bare Demo of IEEEtran.cls for IEEE Journals}
%



\maketitle

\begin{abstract}
Smart Manufacturing refers to optimization techniques that are implemented in production operations by utilizing advanced analytics approaches. With the widespread increase in deploying Industrial Internet of Things (IIoT) sensors in manufacturing processes, there is a progressive need for optimal and effective approaches to data management. Embracing Machine Learning and Artificial Intelligence to take advantage of manufacturing data can lead to efficient and intelligent automation. In this paper, we conduct a comprehensive analysis based on Evolutionary Computing and Deep Learning algorithms toward making semiconductor manufacturing smart. We propose a dynamic algorithm for gaining useful insights about semiconductor manufacturing processes and to address various challenges. We elaborate on the utilization of a Genetic Algorithm and Neural Network to propose an intelligent feature selection algorithm. Our objective is to provide an advanced solution for controlling manufacturing processes and to gain perspective on various dimensions that enable manufacturers to access effective predictive technologies.

\makeatletter{\renewcommand*{\@makefnmark}{}
\footnotetext{}\makeatother}

\end{abstract}


%
\IEEEpeerreviewmaketitle

\section{Introduction}\label{section.intro}
%
%
%
%

\IEEEPARstart{O}{ver} recent decades, the manufacturing industry witnessed tremendous advances in the form of four major paradigm shifts. In the latest industrial revolution, Industry 4.0, manufacturing has embraced the Industrial Internet of Things (IIoT) \cite{Jeschke,Lojka, Ghahramani2020Urban} and Machine Learning (ML) to enable machinery to boost performance through self-optimization \cite{Joze-Tavcar,Bo-Chen,Peter-Palensky,Yang-Liu,Fortino}. Employing computer control over manufacturing phases can make industry processes smart. Broadly speaking, Smart Manufacturing (SM) can be defined as a data-driven approach that leverages IoT devices and various monitoring sensors. Deploying modern technologies, e.g., IoT coupled with Cloud Computing, in manufacturing, provides access to valuable data at different levels, i.e., manufacturing enterprise, manufacturing equipment, and manufacturing processes. With the prodigious amount of manufacturing data at hand, Computational Intelligence (CI) enables us to transform data into real-time manufacturing insights. Manufacturing, then, can be controlled by leading-edge CI and Artificial Intelligence (AI), and tasks are modelled based on experimental observations, to enhance productivity while reducing costs.

\begin{figure}
  \includegraphics[width=\linewidth]{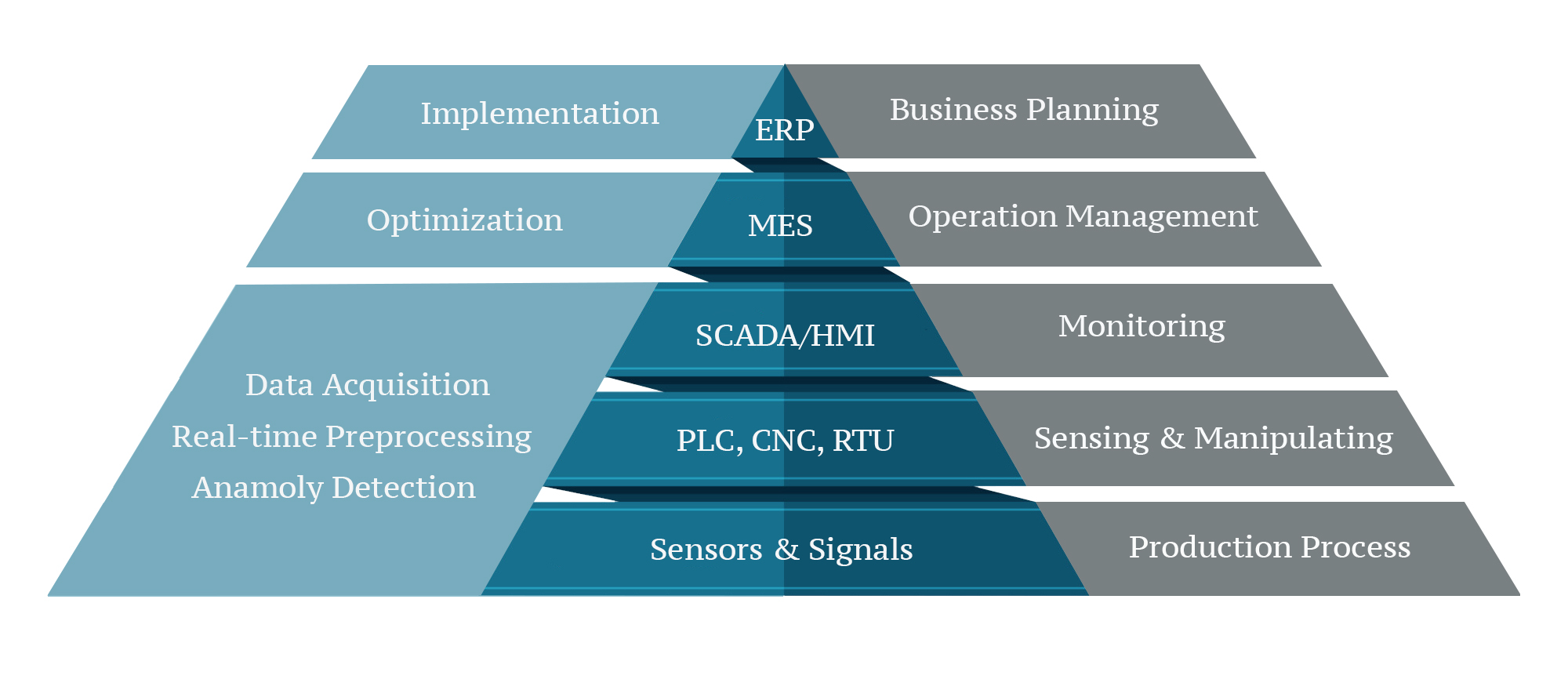}
  \caption{Different levels of automation and their corresponding data analytics (ERP=Enterprise Resource Planning; MES=Manufacturing Execution Systems; SCADA=Supervisory Control and Data Acquisition; HMI=Human Machine Interface; PLC=Programmable Logic Controller; CNC=Computer Numerical Control; RTU=Remote Terminal Unit).}
  \label{pyramid}
\end{figure}

Cost-effective and sustainable manufacturing has become the focus of academia and industry. In doing so, it is of so great importance to identify which factors play a pivotal role in process outcomes. An integrated model based on manufacturing processes and data analytics is demonstrated in Fig. \ref{pyramid}. The model has been divided into different layers and can be considered as a Computer-Integrated Manufacturing (CIM) model from which computational intelligence can take control of the entire production processes. At the Business Planning level, all decisions regarding the end product are made. The operational decisions related to optimizing processes are managed in the Operation Management level. On the Monitoring Level, different sensor-based monitoring approaches, e.g., anomaly detection methods, are employed. Finally, data acquisition and real-time processing are performed at the Production Process level and Sensing level, respectively.

The approach implemented in this work aims to mitigate cost and production risks and promoting sustainable development of semiconductor manufacturing. Moving towards an optimal system, i.e., one that is adaptive and intelligent, is not a trivial task; however, embedding intelligent algorithms in automation and semiconductor production could be beneficial for both reducing cost and enhancing quality of products. The main focus of smart manufacturing studies is on the product life-cycle management, manufacturing process management, industry-specific communication protocols, and manufacturing strategies. Recent advances in technology-based solutions, e.g., IoT, cloud/fog computing, and big data, can expedite and simplify the production process and make new development of manufacturing possible \cite{Ghahramani-1,Ghahramani-2,Ghahramani-3,Ghahramani-4,D-Bertsekas,Ghahramani2018Spatial,Ghahramani2020Intelligent}. These advances should drive the evolution of manufacturing architectures into integrated networks of automation devices and enable the smart characteristics of being self-adaptive, self-sensing, and self-organizing. Providing such solutions includes addressing several challenges, e.g., data volume, data quality and data merging.

Traditional fault detection and diagnosis systems interpret sensory signals as single values \cite{Zhiwei-Gao1}. Then, these values are fed into a model to verify product status. The main drawback of this approach is that it fails to determine the most important features/operations involved in semiconductor production and may result in the loss of sensory data. Moreover, sensory data might consist of noise, outliers, and missing values and can be characterized by heterogeneous structures. To address these concerns, our goal is to propose an intelligent and dynamic algorithm consisting of a feature extraction mode. Generally speaking, predicting the quality of products is an imbalanced classification problem and semiconductor manufacturing is not an exception. To be specific, the dataset is imbalanced because the defective rate in manufacturing processes is quite low in practice. To address this potential issue, a proper imbalanced technique needs to be taken into account to improve model performance. Such implementation is discussed in the following sections. Moreover, we propose an integrated algorithm to solve a multiobjective problem based on a Artificial Neural Network (ANN) and Genetic Algorithm (GA), to establish a fault diagnosis solution by extracting the most relevant features and then using these features as an input for classifiers. It should be mentioned that Multiobjective Evolutionary Algorithms (MOEA) is divided into different categories, i.e., decomposition-based and dominance-based methods. In this work, a decomposition method (weighted sum technique) based on a binary GA and ANN is proposed. This approach is practicable for all kinds of manufacturing analyses in the context of feature extraction/selection, dynamic optimization, and fault detection. Specifically, we investigate the following:

\begin{enumerate}
\item[{1)}] how a hybrid model based on an Evolutionary algorithm combined with ANN can be proposed to model nonlinearity;
\item[{2)}] how to integrate the capabilities of ML combined with AI to implement a highly flexible and personalized smart manufacturing environment;
\item[{3)}] whether combining ML with AI can outperform the traditional methods;
\end{enumerate}

Given the extracted features, various classification methods are tested and the one with the minimum classification error rate are selected. Also, a comparison between the proposed solution and traditional methods is presented. The integrated approach is shown to outperform the others in terms of the accuracy and performance of a manufacturing system. This model can be also useful for fault detection without requiring specialized knowledge. In this implementation, we have encountered several issues, e.g., handling imbalanced data, exploration, and exploitation in an optimization process. To address such concerns, various scenarios are discussed throughout the paper.

\subsection*{Research Methodology and Contributions}
Modern embedded systems, an emerging area of ML, AI, and IoT, can be a promising solution for efficient, cost-effective manufacturing production. Semiconductor manufacturing is a highly interdisciplinary process, complex and costly including various phases. Failures during the manufacturing phases result in faulty products. Hence, detecting the causes of failures is crucial for effective policy-making and is a challenging task in the Business Planning stage as demonstrated in Fig. \ref{pyramid}. This can be achieved by fully exploring production phases and extracting relevant manufacturing features involved in the production. Therefore, fault detection and feature extraction are of much importance. Accordingly, we deal with implementing a model for feature extraction and classification in semiconductor manufacturing. The solution involves developing a model for monitoring processes based on ML and AI algorithms to enhance the overall output of manufacturing processes by extracting the most relevant features. Interpreting these features (manufacturing processes), consequently, provides us with the ability to quickly identify the root cause of a defect. One such efficient model contributes to cost reduction and productivity improvement.

While the greatest challenge in this work is a feature extraction/feature selection task, some other data-related issues, such as imbalanced data and outliers, should first be addressed. These data preparation steps aim to transform raw data into meaningful and useful ones that can be used to distinguish data patterns and consequently enable us to implement effective strategies. To solve the imbalanced classification issue, we have adopted a synthetic minority over-sampling algorithm to boost the small number of defective cases and assign a higher cost to the misclassification of defective products than that of normal products. A confidence interval is defined and outliers have been identified based on this measurement and eliminated. Then, the initial set of data is fed into a feature selection algorithm. Feature extraction aims to project high-dimensional data sets into lower-dimensional ones in which relevant features can be preserved. These features, then, are used to distinguish patterns. 

The proposed dynamic feature selection model is based on an integrated algorithm including a meta-heuristic method (GA) and artificial neural network. We have implemented a binary GA to determine the optimal number of features and its relevant cost which are used to create a predictive model. Our goal is a solution with low-cost values in a search process. The cost function has been defined by using a multilayer perceptron that is considered as an embedded part of a feature selection algorithm. GA consists of different phases, i.e. parent selection, crossover, mutation, and creating final population (the selected features) \cite{Alinodehi}. Parent Selection is a crucial part of GA and consists of a finite repetition of different operations, i.e., selection of parent strings, recombination, and mutation. The objective in a reproductive phase is to select cost-efficient chromosomes from the population, which create offsprings for the next generation. To address the exploration and exploitation and to avoid premature convergence, we have proposed a selection scheme by combining different crossover operations. The mentioned issue is heavily related to the loss of diversity. The proposed solution also eliminates the cost scaling issue and adjusts the selection pressure throughout the selection phase. We adjust the balance between exploration and exploitation by recombining crossover operators with adjustment of their probabilities. A discussion for determining exploration and exploitation rate is presented in the following sections. Consequently, offsprings are created by adjusting such probabilities throughout the mating pool by establishing a hybrid roulette-tournament pick operator. Selected features are fed to a predictive model to determine fault status. It is worth mentioning that the algorithm considers two major conflicting objectives: minimizing the number of features and maximizing the classification performance. Consequently, the result of the proposed model are compared with traditional approaches. The experiments have verified the effectiveness and efficiency of our approach as compared to those in the literature. In summary, the overall objective is to propose an AI-based multi-objective feature selection method together with an efficient classification algorithm to scrutinise manufacturing processes.

The remainder of this paper is organized as follows: some related work about manufacturing processes, feature extraction and application of AI is described in Section \ref{section.RelatedWorks}; a preprocessing procedure is discussed in Section \ref{section.preprocessing}; the proposed approach with its associated discussions is given in Section \ref{model}; the experimental settings and the classification results are shown in Section \ref{result}; and the future work and conclusions are presented in Section \ref{section.conclusion}.

\section{Related Work}\label{section.RelatedWorks}

Recently, the rapid evolution of high-throughput technologies has resulted in the exponential growth of manufacturing data \cite{Jiafu-Wan}. Since traditional approaches toward data management are impractical due to high dimensionality, proposing an effective and efficient data management strategy has become crucial. To do so, ML can help develop strategies to automatically identify patterns from high dimensional datasets. The key to leveraging manufacturing data lies in constant monitoring of processes, which can be associated with different issues, e.g., noisy signals. Dimensionality reduction and feature selection/extraction methods, e.g. Principal Component Analysis (PCA), Linear Discriminant Analysis (LDA), and Canonical Correlation Analysis (CCA), play a critical role in dealing with noise and redundant features and must be considered as a preprocessing stage of manufacturing data analysis, which leads to better insights and robust decisions \cite{Semyon-Meerkov}. Some previous manufacturing fault detection studies have focused on utilizing the mentioned techniques for extracting the most relevant features and classification. Feature selection methods can be divided into three main categories, i.e., filter, wrapper, and embedded methods. The filter methods act based on ranking the features. In wrapper methods, features are selected based on the performance of predictors. Finally, embedded methods include variable selection as part of the training process without splitting the data into training and testing sets.

In \cite{He-}, the authors have utilized PCA to extract features to decrease the computational cost and complexity. Given the extracted features, they have implemented a classification algorithm to infer whether a semiconductor device is a defective or normal sample. To that end, they have adopted a k-nearest neighbors (KNN) classification method. Cherry et al. have conducted another model based on a multiway PCA (MPCA) to monitor stream data \cite{Cherry-}. A decision tree algorithm has been developed in \cite{He-2-} to explore various types of defective devices. A KNN method has been utilized in \cite{He-3-}, and Euclidean distance has been considered to measure similarities among features. Verdier et al. have improved the performance of a KNN algorithm tailored for fault detection in semiconductor manufacturing by defining similarity measurement based on Mahalanobis distance \cite{Verdier-3}. A support vector machine (SVM) is used to detect semiconductor failures in \cite{Baly-3}. The authors have developed their approach based on an RBF kernel to address the high dimensionality issue. In \cite{Kwak-3}, an incremental clustering method is adopted for fault detection. A Bayesian model has been proposed to infer a manufacturing process. The authors have considered the root causes of manufacturing problems. However, their approach heavily relies on an expert's knowledge regarding the related field. Zheng et al. have proposed a convolution neural network \cite{Zheng-3}. They have decomposed multivariate time-series datasets into univariate ones. Then, features have been extracted and an MLP-based method has been implemented for data classification. Lee et al. have compared the performance of different fault detection models, including feature extraction algorithms and classification approaches \cite{Lee-3}. They have revealed that developing an algorithm based on features that are not suitable for a specific model can deteriorate the performance of classifiers significantly. Therefore, it is desirable to consider both feature extraction and classification stages simultaneously to maximize a model's performance.

Most studies in the literature have focused on using PCA and KNN algorithms for manufacturing data classification. However, PCA-based approaches project features to another space based on a linear combination of original features. Therefore they cannot be interpreted in the original feature space \cite{Zhang-pca}. Moreover, most of the PCA-related work has considered linear PCA, which is not efficient in exploring non-linear patterns. Although these techniques try to cover maximum variance among manufacturing variables, inappropriate selection of parameters, e.g., principal components, may result in great data loss. KNN is a memory-based classifier. Hence, in cases of high dimensional data sets, its performance degrades dramatically with data size. To overcome the mentioned concerns, an efficient global search method (e.g., Evolutionary computation (EC) techniques) should be considered to better address feature selection problems \cite{Xue2016Survey}. These techniques are well-known for their global search ability. Derrac et al. \cite{Derrac-first} have proposed a cooperative co-evolutionary algorithm for feature selection based on a GA. The proposed method addresses feature selection task in a single process. However, it should be mentioned that EC algorithms are stochastic methods, which may produce different solutions when using different starting points. Therefore, the proposed model suffers an instability issue. Zamalloa et al. \cite{Zamalloa-Feature} have utilized a GA-based method to rank features. Consequently, features have been selected given the rank orders. A potential drawback of this work is that the proposed method might lead to data loss. Moreover, this solution has not considered the correlation among features. 

To address the mentioned concerns, we have proposed our solution based on a dynamic feature selection method consisting of different modes to provide information on the variables that are crucial for fault diagnosis. To that end, we have integrated ANN into our model in order to examine nonlinear relationship among features. Advanced computing and AI can provide manufacturing with a higher degree of intelligence and low-cost sensing and improve efficiency \cite{Dong-3}. The process of conducting intelligent manufacturing can be regarded in two ways. Firstly, the manufacturing industry has become a  great contributor to the service industry and secondly the lines between the cyber and physical systems are becoming blurred. Hence, architectural approaches like service-oriented architectures (Cloud manufacturing) can be taken into account in manufacturing modes and systems. In such distributed and heterogeneous systems, manufacturing resources can be aggregated based on an efficient service-oriented manufacturing model and processed/monitored in an effective way. Application of those solutions can pave the way for large-scale analysis and leads to high productivity. Developing a successful model includes various steps, e.g., data cleansing and data transformation, to reveal insights. As the quality of data affects the analysis, it is essential to employ a data preprocessing procedure. Such discussion is demonstrated next.

\section{Data preprocessing} \label{section.preprocessing}
\indent The data set used in this work is obtained from a semiconductor factory, SECOM (Semiconductor Manufacturing) dataset. It consists of various operation observations, i.e., wafer fabrication production data, including 590 features (operation measurements). The target feature is binomial (Failure and Success), referring to the production status, and encoded as $0$ and $1$. The first step in data analysis is data cleansing to address a variety of data quality issues, e.g., noise, outliers, inconsistency, and missing values. We have dealt with missing value and noise resulting from inexact data collection. These can negatively affect a later processes. Outlier labelling methods and T-squared statistics ($T^2$) have been utilized. Any observation beyond the interval has been eliminated. 

\subsection{Outlier Detection}
Suppose that, $F=\{f_{1}, f_{2}, \ldots, f_{m}\}$ denotes the feature set and $L=\{Failure, Success\}$ the label set, where $m$ is the number of features.

Matrix $X \in \mathbf{R}^{n \times m}$ can be defined as:

 \begin{equation}
X=\begin{bmatrix}
         X_{1} \\   X_{2} \\ \vdots \\ X_{n}
   \end{bmatrix}
=\begin{bmatrix}
         x_{11} & x_{12} & \cdots  & x_{1m} \\   
         x_{21} & x_{22} & \cdots  & x_{2m} \\ 
         \vdots  & \vdots  & \ddots & \vdots  \\ 
         x_{n1} & x_{n2} & \cdots  & x_{nm} 
   \end{bmatrix}
  \end{equation}
where $\mathbf{R}$ is the real number set, $X_i$ (the $i^{\text{th}}$ observation) is defined as an \textit{m}-tuple ($m$ is the number of features), containing all features, and $n$ is the number of observations.

The label feature $Y$ is as follows:

 \begin{equation}
Y=\begin{bmatrix}
         y_{1} ,   y_{2} , \ldots, y_{n}
        \end{bmatrix}^\top
  \end{equation}
where $y_{i}$ is the corresponding label (Success or Failure) for the $i^{\text{th}}$ observation ($X_i$) and $\top$ is a transpose operator.

We utilize the Mahalanobis distance of each observation ($X_i$) from the mean, i.e.,

\begin{equation}
D = (X_i - \bar X) {S} ^{-1} (X_i - \bar X)^\top
\end{equation}
where $S^{-1}$ is the inverse of the $m \times m$ variance-covariance matrix (Scatter matrix) and $\bar X = \frac {1}{n} \sum_{i=1}^{n} X_{i} $. The Mahalanobis distance and the values are chi-square distributed. The variance-covariance matrix can be calculated as:

\begin{equation}
S = \frac{1}{n-1}\sum_{i=1}^{n} (X_i - \bar X)^T(X_i - \bar X)
\end{equation}
where $\alpha$ is the confidence interval. Given $\alpha$, if $(X_i - \bar X) {S} ^{-1} (X_i - \bar X)^\top > \alpha ^ 2$, $X_i$ is treated as an outlier and eliminated. For this purpose, a quantile of the ${\chi}^2$ distribution (e.g., the $97.5 \%$ quantile) is considered.

\subsection{Handling an Imbalanced Data set}
The observations that have been labeled as \textit{Failure} are relatively rare (104 cases) as compared to the \textit{Success} class. Hence, we face an imbalanced classification issue. In other words, \textit{Success class} (the majority) outnumbered \textit{Failure class} (the minority) and both classes do not make up an equal portion of our data set. Two distinctive approaches can be considered to deal with this issue: 1) skew-insensitive methods and 2) re-sampling methods. The first category addresses the problem by assigning a cost to the training data set while the second one adjusts the original data set such that a more balanced class distribution is achieved. Re-sampling methods have become standard approaches and have been dominantly utilized recently \cite{Shuo-Wang,Liu-Mengchu,Q-Kang}. They can be classified into different categories, e.g., sampling strategies, wrapper approaches and ensemble-based methods. Implementing a proper method is crucial, otherwise it can be problematic, e.g., data loss and overfitting, and can result in a poor outcome. Our goal in this phase is to relatively balance class distribution. To do so, we have utilized a synthetic minority over-sampling technique. There are various over-sampling algorithms, such as SMOTE, Borderline-SMOTE, and Safe-Level-SMOTE, just to name a few. The mentioned methods create synthetic samples based on the nearest neighbour approach and can be negatively impacted by the over-generalization issue. To overcome these problems in this work, a density-based SMOTE \cite{Bunkhumpornpat-1}-\cite{Xuesong-Zhang} technique is utilized and by synthetically adding \textit{Failure class} instances we make the distribution more balanced. It is an over-sampling method in which the Failure class is over-sampled by generating its synthetic instances.

\subsection{Feature Selection}
    As stated, the data set consists of nearly 600 features. Data sets with high dimensions can cause serious challenges such as overfitting in learning processes, known as the curse of dimensionality. To address these challenges the dimensionality needs to be reduced and different approaches have been proposed in the literature. Generally speaking, dimensionality reduction can be considered as an approach to eliminate redundant (or noisy) features. It can be divided into two categories, \textit{feature extraction} and \textit{feature selection}. The former refers to those methods (PCA and LDA) that map original features to a new feature space with lower dimensionality while the latter aims to select a subset of features such that the trained model (based on the selected features) minimizes redundancy and maximizes relevance to the target feature. PCA (a classic approach to dimensionality reduction), Multidimensional Scaling, and Independent Component Analysis (ICA) all suffer from a global linearity issue. To address the mentioned shortcoming, nonlinear techniques have been proposed: kernel PCA, Laplacian eigenmaps and semidefinite embedding. Since reconstructing observations (after the projection phase) in these nonlinear methods is not a trivial task, finding the corresponding pattern is sometimes impractical. In a feature extraction approach, observations are projected into another space where there is no physical meaning between newly generated features and the original ones. Hence, feature selection methods are superior in terms of readability and interpretability in this sense. Therefore, to avoid complexity and uncertainty that feature extraction techniques bring, a feature selection approach has been opted for in this work. To this end, we have proposed an integrated approach, consisting of a \textit{metaheuristic} algorithm (GA) and an Artificial Neural Network. GA is a heuristic search method and inspired by Charles Darwin's theory of natural evolution. Since selecting features can be considered as a binary problem, we have developed our model based on binary GA that treated candidate features (chromosomes in GA terminology) as bit-strings.

\begin{figure*}
  \includegraphics[width=\linewidth]{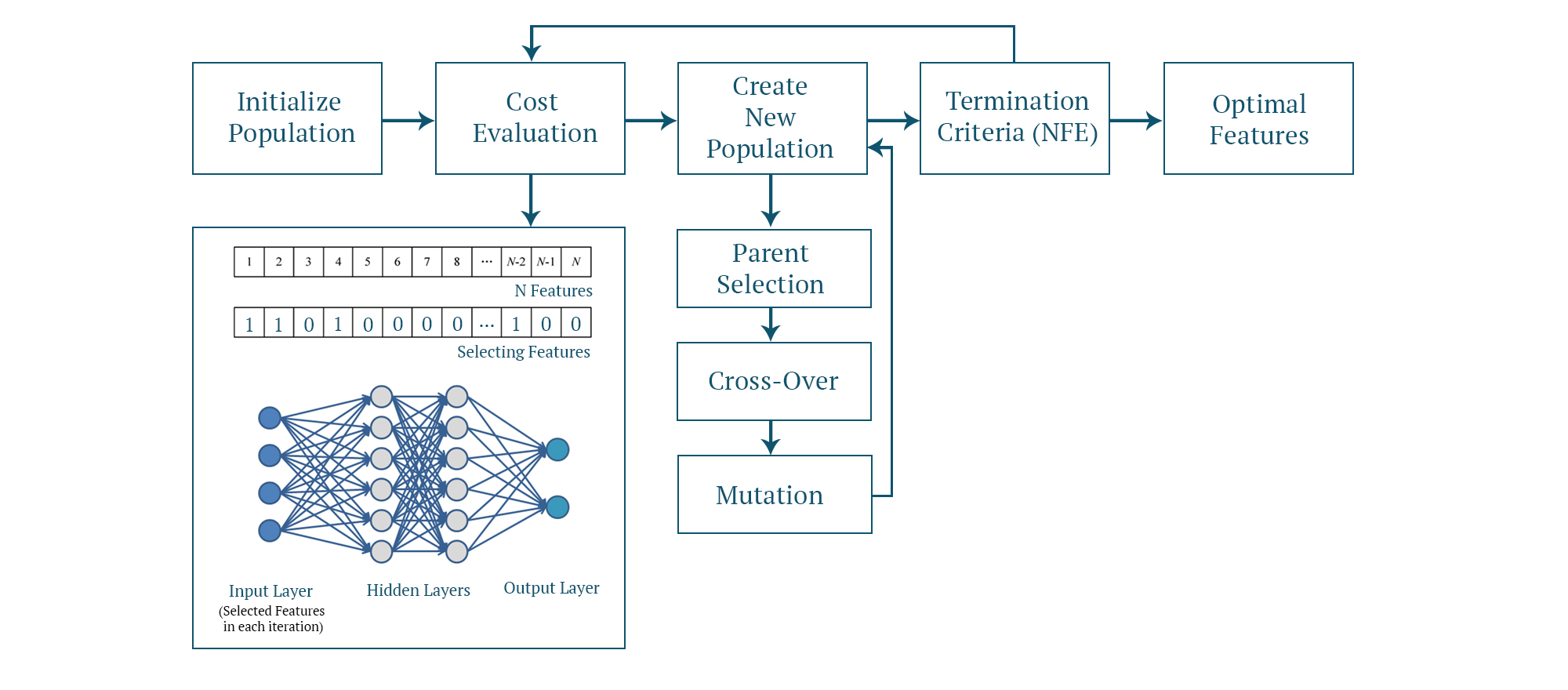}
  \caption{Feature Selection Model Using Artificial Neural Network and Genetic Algorithm}
  \label{featureSelection}
\end{figure*}

GA relies on a population of individuals to explore a search space. Each individual is a set of chromosomes, encoded as strings of $0$ (if the corresponding feature is not selected) and $1$ (if the feature is selected). GA utilizes an initial population and some genetic operators, e.g., crossover and mutation, to generate a new generation by recombining a population's chromosomes. Then, fitter individuals are selected according to a cost-function (objective-function) in a reproduction phase. GA maintains its effectiveness from two sources: \textit{Exploration} and \textit{Exploitation}. The former can be considered as a process of exploring a search space (by genetic search operators, e.g. crossover operation), while the latter is the process of employing a mutation operator and modifying offsprings' chromosomes. A balance between the mentioned abilities (Exploration and Exploitation) should be maintained. To that end, beneficial aspects of existing solutions (individuals with lower costs) should be exploited. Moreover, exploring the feature space in order to find an optimal solution (optimal features) is crucial. While a crossover operation is the main search operator, a mutation operator is employed to avoid premature convergence. The level of exploration/exploitation can be controlled by selection processes, e.g. selection pressure parameter. Selecting an appropriate pressure measurement ($\beta$ in this work) can maintain a balance between exploration and exploitation. Such discussion is provided in Section \ref{model}. Parameter $\beta$ has been used in the parent selection stage and candidate individuals have been taken into account in the generation production. This operation, iteratively, has been repeated until the termination criteria (number of iterations or number of function evaluations (NFE)) are met. The best individual (the one with the minimum cost) is selected and in this way optimal features are then identified. Fig. \ref{featureSelection} displays our proposed feature selection model.

\section{Feature Selection Model}\label{model}
\begin{figure}
  \includegraphics[width=\linewidth]{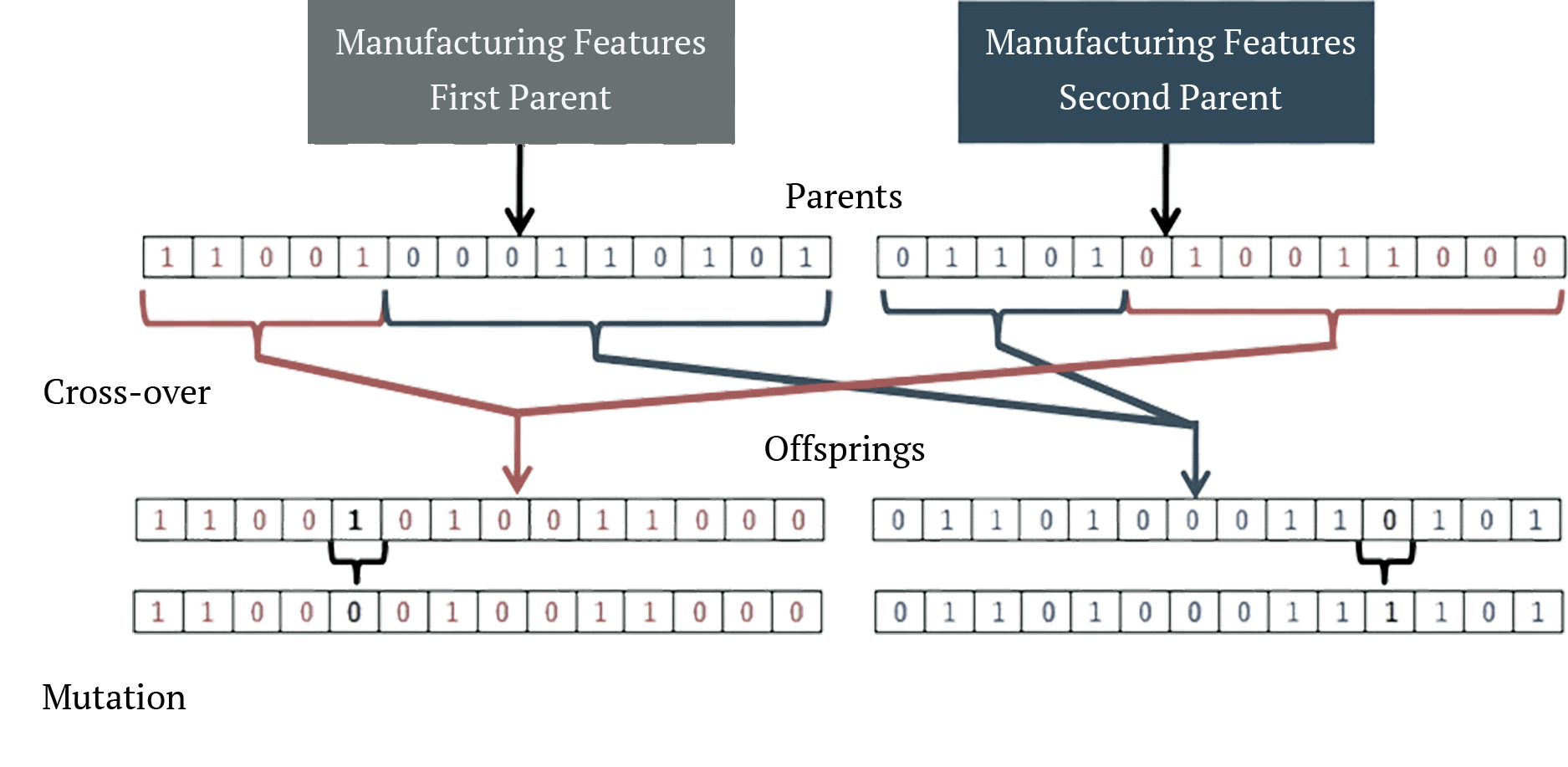}
  \caption{Reproduction phase.}
  \label{Reproduction}
\end{figure}

As mentioned earlier, our objective is to modify the output of each iteration (a subset of features) by searching feature space and finding proper values for the input features such that the measured cost is minimized. Given Fig. \ref{featureSelection}, our proposed feature selection model consists of different phases. It starts with defining an initial population, i.e., individuals including $m$-dimensional chromosomes.

  \begin{equation*} 
  \zeta= (v_{1}, v_{1}, \ldots, v_{m})
  \end{equation*}
where $v_{i}$ is either $1$ or $0$, and corresponds to the status of the $i^{\text{th}}$ variable (feature), selected or not. While some individuals are admitted to the new generation unchanged, others may be subject to some genetic operators (crossover and mutation). The cost related to each individual is evaluated by the ANN in the second phase to be presented in Section \ref{model}. These costs are utilized (in the parent selection phase) to determine which offsprings are used to create a new generation. The objective is to select two individuals (with lower costs) from the population such that newly created offsprings would inherit such patterns from their parents. Generally speaking, there are several approaches to select parents such as Random Selection, Rank Selection, Stochastic Universal Sampling (SUS), Tournament Selection, and Boltzmann Selection. There is no selection pressure parameter in the Random Selection method, and hence it is usually avoided. Rank Selection and SUS suffer from premature convergence and applying such approaches may easily result in a local optimum. To avoid the mentioned situation and maintain good diversity we have employed the Boltzmann Selection method which is inspired by Simulated Annealing. The probability of an individual being selected is calculated according to the below Boltzmann probability:

   \begin{equation*} 
  p_{(i)}= \frac{e^{-\beta J_{i}}}{\sum_{k=1}^{\eta_{p}} e^{-\beta J_{k}}}
  \end{equation*}
where $\eta_{p}$ is the size of the initial population and $J$ is the defined cost function. $\beta$ is the selection pressure. It is clear that parents are selected based on probabilities which are proportional to the costs measured in the prior phase. This means that individuals with a lower cost are more likely to be chosen than ones with a greater cost. It should be mentioned that we have selected the $\beta$ parameter such that $\sum_{i\in H}^{} p_{(i)}=0.7$, where $H$ is the set of half of the best individuals (population is sorted according to their cost values and $\eta_{p}/2$ of them are selected). Consequently, the Roulette Wheel method is utilized for sampling (selecting parents using stochastic sampling with replacement based on Boltzmann probability function). A circular wheel is considered and divided into $\eta_{p}$ pies, each of which is proportional to the cost values. The wheel is spun and the individual related to the pie on which it stops is then selected. We have repeated this procedure until our predefined number of parents are selected. In this way, individuals with the largest cost value have the minimal chance to be selected. Parents are selected according to the weighted slots, cross-over operations are then applied to them. On this basis, the chromosomes of selected parents are combined to create new offspring. As demonstrated in Fig. \ref{Reproduction}, a random portion of the first individual is swapped with a random portion of the second one. In this process the chromosomes combination can be carried out in different ways e.g. single-point, double-point, or uniform crossover. In single-point cross-over, one random position in the array of bits is selected and exchanging then takes place, while in double-point method, two positions are chosen and chromosomes are swapped. In uniform cross-over, parents' chromosomes are selected for random exchange. Parents contribute to creating new offspring based on a bit string known as the cross-over mask. Let $\xi$ be the predefined cross-over mask, e.g., $\xi= \{1,1,0,0,0,1, \ldots, 0,0,1\}$. As discussed earlier, after the initial population is created, the parent selection operation should be conducted in the reproduction phase. Our goal is to select individuals from those with minimum costs in the population. Consequently, parents are selected to create offspring for the next generation. The cost function and the way we have integrated ANN to calculate this measurement is next described. 

\subsection{Cost Function and MLP}
Our objective in the feature selection phase is to explore a hypothesis space, find the optimal number of features, and consequently reduce the dimensionality. In other words, we are looking for a subset of the original dataset, $J: X^\prime \subseteq X \to \mathbf{R}$, such that two criteria are to be met. The cost function obtains different subset of features and target values as the input and the corresponding costs are calculated. Given the conventions adopted earlier, let $F$ be the original feature set with the cardinality of $|X|= m$. Now, let $J(X^\prime)$ be an evaluation measure to be optimized given the below criteria:

\begin{itemize}
  \item Set $|X^\prime|=k<m$. Find $X^\prime \subset X$, such that $J(X^\prime)$ is maximized. It is equal to minimizing the Mean Squared Error: 
  
  \begin{equation*} 
  \epsilon= \frac{1}{n} \sum_{i=1}^{n} (Y_{i}-J_{i})^2
  \end{equation*}
  
  \item Find optimal features (in the case of both the number of features and discrimination) while minimizing $|X^\prime|=n_{j}$.
\end{itemize}

It should be mentioned that we are facing a multi-objective optimization problem \cite{Gupta-3}. We define our objective function with weights as:

  \begin{equation}  
   J = \epsilon \times (1+(\Omega \times |X^\prime|))
  \end{equation}
where $|X^\prime|$ is the dimension of selected features in each iteration, and $\Omega$ can be considered as a cost parameter for choosing new features. If $\Omega=0$, all features are selected, while a large number to $\Omega$ results in no feature being selected. This parameter is a trade-off between relevancy and redundancy and must be designated carefully. As stated, our objective is to minimize objective function $J$. In doing so, we have integrated Artificial Neural Network (ANN) and GA. GA gets the defined cost function (i.e., Feature-Selection-Cost, $J$) as the input and employs the ANN to calculate cost values. Iteratively, different individuals (bit strings) consist of $0$ and $1$ (where $1$ refers to a feature being selected and $0$ refers to is not being selected) are generated and evaluated by GA's operations. Multilayer Perceptron (MLP) is utilized to calculate $\epsilon$ (in each iteration). The Multi-layer perceptrons with Levenberg-Marquardt training algorithm is used (since it converges faster and more accurately towards our problem) and consisting of two layers (15 neurons in the hidden layer) of adaptive weights with full connectivity among neurons in the input and hidden layers. All costs are calculated and the best features are selected such that the corresponding cost is minimized. To summarize the procedures, the pseudo code of the feature selection model is presented in Algorithm \ref{alg:GA}. 

\section{Analysis Procedures} \label{result}
As mentioned earlier, in this work, we deal with a classification problem with a relatively large number of variables. It has been widely discussed \cite{Chenping-Hou} that irrelevant variables may deteriorate the performance of algorithms. The application of feature extraction/selection methods can make it possible to choose a subset of features possible and thus helps achieve reliable performance. Most studies in the literature have considered feature selection as a single objective problem while our solution is based on a multi-objective approach. In this section, different approaches, i.e., conventional feature extraction methods and the model proposed in this work, are compared. Our objective is to demonstrate that an intelligent algorithm can outperform the results of other competing classification methods.

\subsection{Feature Extraction Methods}
Different scenarios in the context of feature extractions are available to remove irrelevant features. All solutions have been considered as a pre-processing task in order to increase the learning accuracy. These conventional methods can be categorized into the filter, wrapper, embedded and hybrid techniques. Filter methods are divided into univariate and multivariate layers. The relevance of features is evaluated based on ranking techniques. Wrapper methods, e.g. sequential selection and heuristic search algorithms, are basically a search algorithm and relevant features are selected by training and testing a classification model. Embedded methods are performed based on dependencies among features. Finally, the hybrid method is based on a combination of other approaches and consists of different phases. These methods have some serious drawbacks which can make their results unrealistic. Filter methods do not consider the features' dependencies and the relationship between independent and dependent features. There is a high risk of an overfitting problem in the wrapper approach. Embedded methods are more of a local discrimination approach than a global one and the hybrid methods are computationally expensive. Next we compare the results of the models implemented in this work.

\subsection{Results}
The experiment has been conducted on a computer with quad-core Intel i9-7900X CPU 8 GHz processor and 32 GB memory. It was equipped with a NVIDIA GeForce GTX 1080 GPU and 8GB memory. The parallel algorithm has been implemented by CUDA programming.

\begin{figure}
  \includegraphics[width=\linewidth]{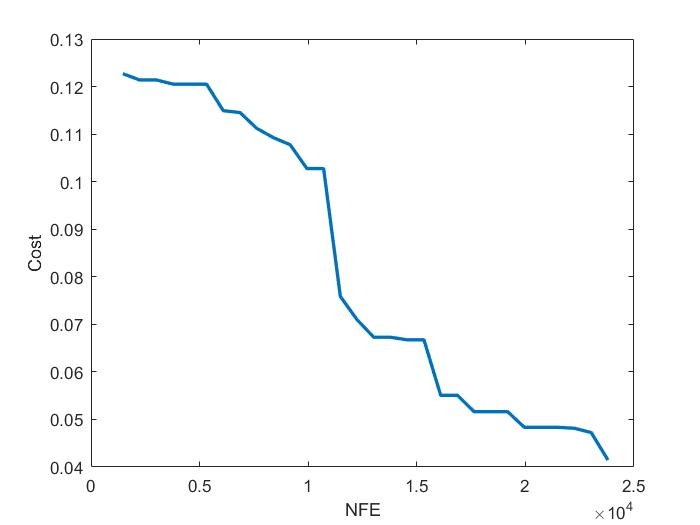}
  \caption{Cost function value versus Number of Function Evaluations (NFE)}
  \label{fig:NFE}
\end{figure}

The proposed algorithm for feature selection is based on an adaptive and dynamic GA combined with a neural network. Our meta-heuristic method evaluates various subsets of features to optimize our defined cost function whose calculation has been given to a multilayer perceptron. We consider the volume of our data and the number of features and samples for defining the initial population rate. We choose the number of neurons based on a trial and error method. It should be mentioned that we have used the neural network as a cost function and in this context the main objective is to decrease the cost function's values. The algorithm gets the initial solutions (manufacturing operations) and obtains the optimal features after a series of iterative computations (given the termination criteria, e.g., Number of Function Evaluation). Fig. \ref{fig:NFE} displays the cost values in each iteration.

\begin{figure*}
  \includegraphics[width=\linewidth]{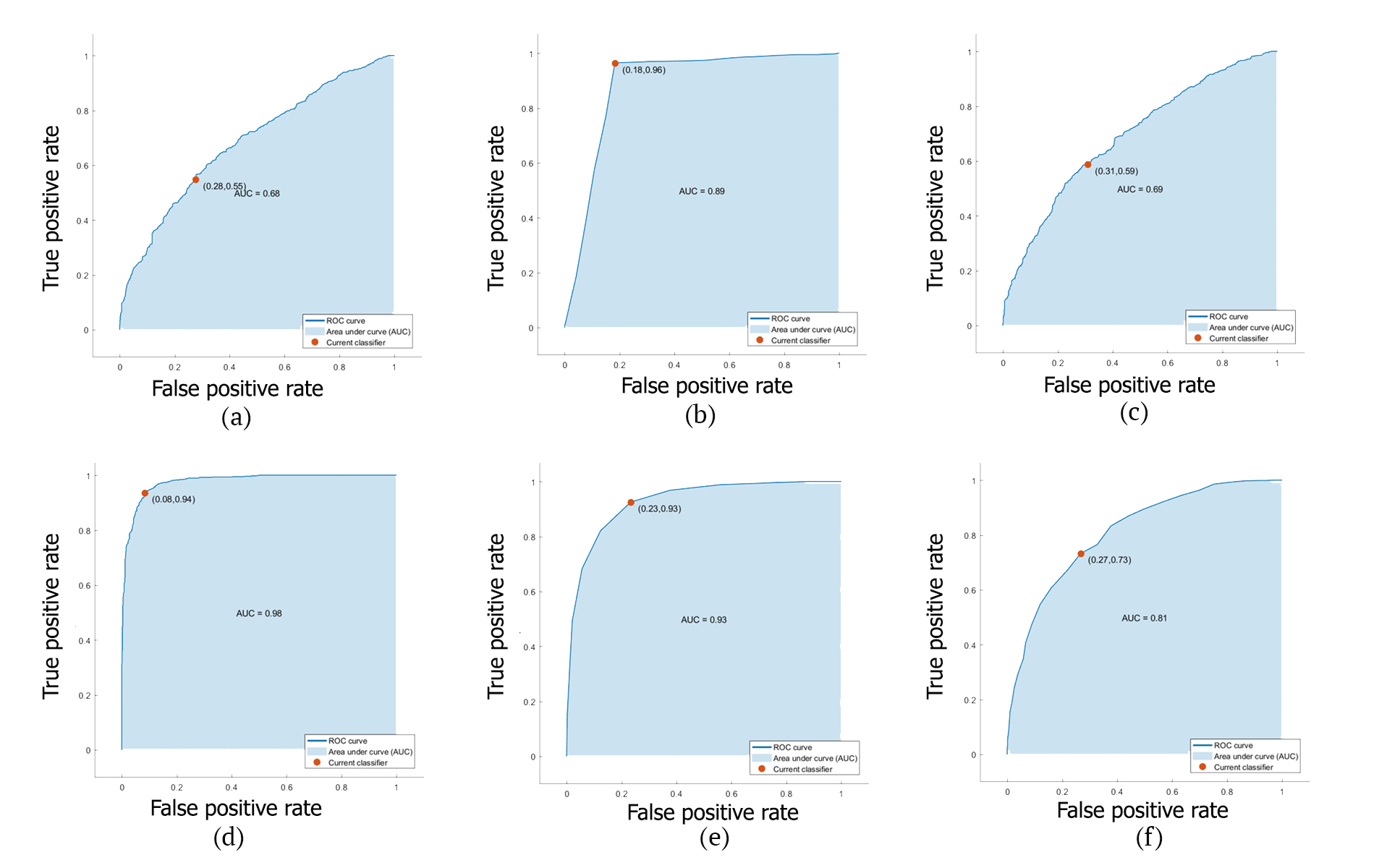}
  \caption{ROC curves for different classification methods: (a) Linear Discriminant, (b) Random Forest, (c) Logistic Regression, (d) Gaussian SVM, (e) k-NN, and (f) SVM with RBF kernel}
  \label{fig:classificatioModels}
\end{figure*}

Finally, we have examined various classification techniques, and the most appropriate one is selected. To do so, different classification models, e.g., Gaussian Support Vector Machine, Random Forest, Linear Discriminant, K-NN, and SVM with RBF kernel, have been tested. The classifiers' performance is evaluated according to their classification accuracies. The ability of each method to accurately predict the correct class is measured and expressed as a percentage. ROC curves are used to determine the predictive performance of the examined classification algorithms. The area under a ROC curve can be considered as an evaluation criterion to select the best classification algorithm. When the area under the curve is approaching $1$, it indicates that the classification has been carried out correctly. Fig. \ref{fig:classificatioModels} shows AUC - ROC curves resulted from implementing different classification methods.

Some statistical results (e.g., the percentage of correct predictions) have been also provided in Table \ref{validity}.  Given the results demonstrated, Gaussian SVM has been selected as the classification model. Some other feature selection methods are also utilized to compare their results with our proposed approach. The discussion regarding it is presented next.

\begin{table*}[!ht]
{\small
  \centering
  \caption{Comparisons of Different Classification Machine Learning Models}
  \label{table:alg}
  \begin{tabular}{llll}
    \toprule
    {\bfseries Classification Method} & {\bfseries Positive Predictive Value (\%) (Success class)}  & {\bfseries False Discovery Rate (\%)(Failure class)}\\
    \midrule
    Linear Discriminant &  64 & 36 \\
    Random Forest &  83  & 4  \\
    Logistic Regression &  63 & 35 \\
    \textbf{Gaussian SVM} &  \textbf{91} & \textbf{6}\\
    Adaptive k-NN &  84 & 16 \\
    SVM with RBF kernel&  71 & 25 \\
    \bottomrule
  \end{tabular} \label{validity}
  }
\end{table*}

\subsection{Conventional Methods}
\begin{figure*}
  \includegraphics[width=\linewidth]{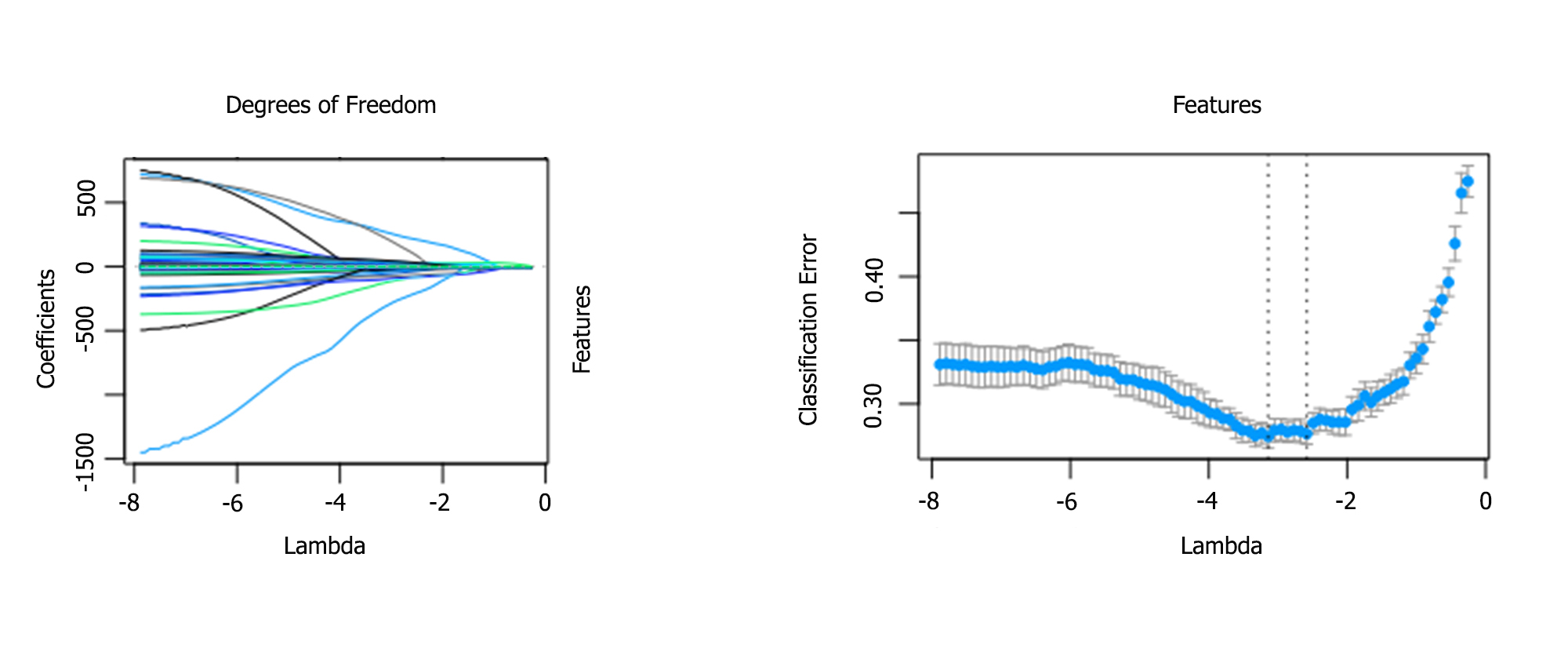}
  \caption{Selecting features based on a conventional method, i.e., Lasso regression. The panels show the lasso coefficient estimates and the curve of the measurements for the degrees of freedom of the Lasso}
  \label{fig:lasso}
\end{figure*}

As discussed in the previous sections, most studies regarding manufacturing data analysis have considered PCA-based approaches which aim to detect the directions of most variation. Together with PCA, we have tested most popular algorithms, e.g., Family-Wise Error Rate (FWE), False Discovery Rate (FDR), Sequential Forward Selection (SFS), Sequential Backward Selection (SBS), Filtration Feature Selection (FFS), Correlation-based Feature Selection (CFS), Lasso Regression and Ensemble methods, for feature extraction \cite{surveyFS}. We have implemented these traditional methods to reduce the dimensionality of our data set and compared the results. To do so, the extracted features are used as the input for the chosen classifier. Fig. \ref{fig:lasso} displays the analysis based on the Lasso Regression method.

\begin{table*}[h]
{\small
  \centering
  \caption{Algorithms Comparisons}
  \label{table:alg}
  \begin{tabular}{lllll}
    \toprule
    {\bfseries Feature extraction method} & {\bfseries Selected Features}  & {\bfseries Accuracy(\%), train data}& {\bfseries Accuracy(\%), test data}\\
    \midrule
    Lasso Regression &  54 & 78.8 & 74.2\\
    PCA &  48  & 87.8 & 79.2 \\
    Univariate Method (FWE) &  15 & 74.3 & 47.8\\
    Controlling False Discovery Rate (FDR) &  12 & 60.7 & 41.5 \\
    Select Percentile &  71 & 84.0 & 81.5 \\
    Sequential Forward Selection &  88 & 81.2 & 72.9 \\
    Sequential Backward Selection & 40 & 78.0  & 72.4 \\
    Filtration Feature Selection & 111 & 73.9  & 43.6 \\
    Correlation-based Feature Selection (CFS) & 92 & 65.2  & 61.0 \\
    Ensemble & 76 & 73.3  & 70.9 \\
    \textbf{Proposed method} & \textbf{36} & \textbf{93.7}  & \textbf{90.2} \\

    \bottomrule
  \end{tabular} \label{validity2}
  }
\end{table*}

\begin{figure*}[h]
  \includegraphics[width=\linewidth]{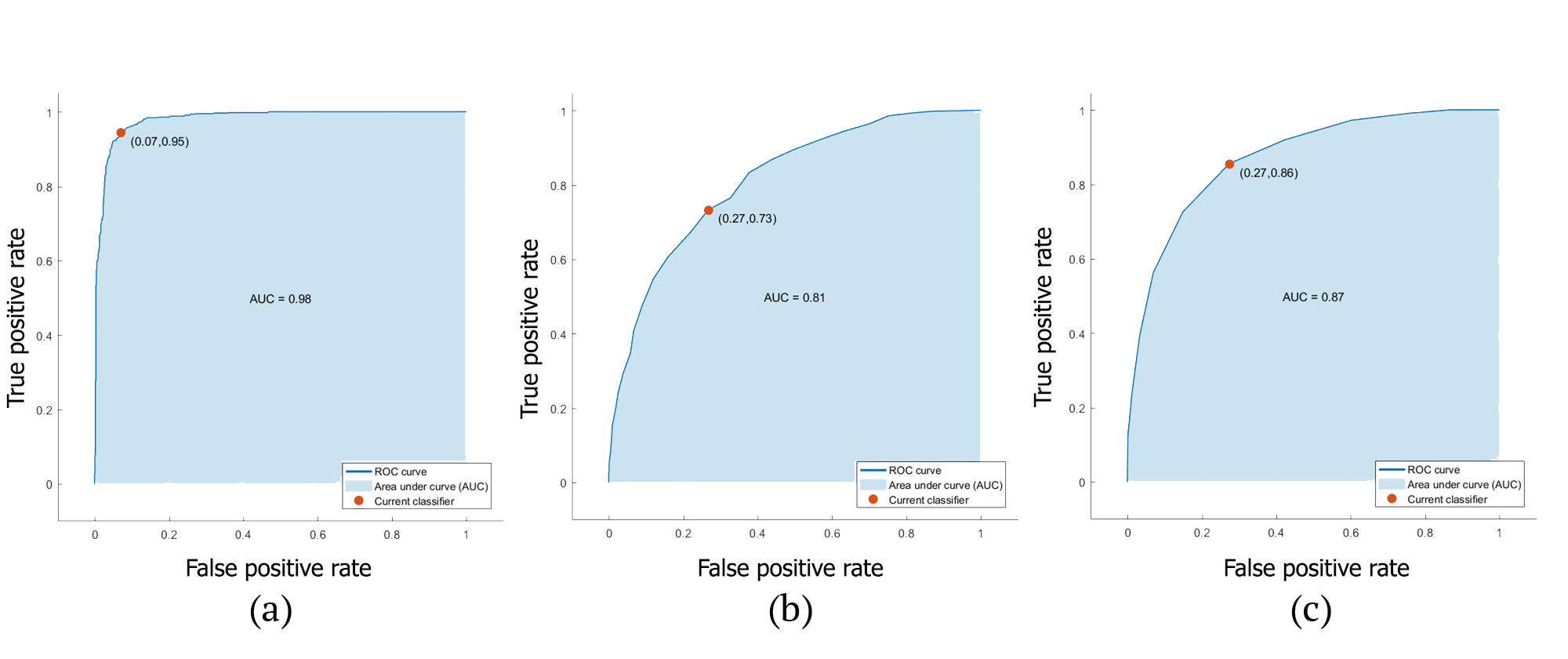}
  \caption{Comparing ROC results from the proposed method (a) vs PCA (b) vs Lasso Regression (c)}
  \label{fig:lassoPCA}
\end{figure*}

The experimental results (Table \ref{validity2}) show that our proposed model is superior over those conventional ones. The corresponding accuracy rate of the proposed model is over 90\%. An ROC comparison between our method and two of the traditional techniques is demonstrated in Fig. \ref{fig:lassoPCA}.

\section{Conclusions and Future Work}\label{section.conclusion}
The goal of manufacturing enterprises is to develop cost-effective and competitive products. Manufacturing Intelligence can significantly improve effectiveness by bridging business and manufacturing models with the help of low-cost sensor data. It aims to achieve a high level of intelligence with the latest appropriate technology-based computing, advanced analytics, and new levels of Internet connectivity. The landscape of Industry 4.0 includes achieving visibility on real-time processes, mutual recognition, and establishing an effective relationship among the workforce, equipment, and products. Most work in the area of manufacturing data analysis are based on PCA-based approaches. They are not able to recognize nonlinear relationships among features and extract complex pattern. To address this concern, we have proposed a dynamic feature selection method based on GA and ANN. We have compared the results achieved in this work with traditional approaches to prove the effectiveness of our proposed solution. As a part of our future work, we plan to consider other MOEAs, e.g., dominance-based algorithms, for solving our optimization problem in a way that both feature selections objective functions are optimized simultaneously. Moreover, we will also compare the current model with other evolutionary algorithms proposed for feature selection.

\appendices
\section{Pseudo-code for the feature selection model}

GA has different parameters and the performance of a GA-based model depends on these parameters. We have discussed how they have been selected throughout this work. Table \ref{tab:Params} reveals the impacts of different parameter setting.

\begin{algorithm*}[!h]
\SetAlgoLined

    \SetKwProg{Fn}{Function}{}{}
    \SetKwInOut{Input}{Input}
    \SetKwInOut{Output}{Output}
    \Input{$ GA (|F|, CostFn)$} 
    \Output{$Individual\gets [\text{Selected Features (a binary vector)}] $}

	{$I^{Max} \gets \text{Maximum number of Iterations}$}\\
	{$ \theta \gets \text{Crossover rate}$},
	{$\mu \gets \text{Mutation rate}$}\\
	{$\eta_{p} \gets \text{Size of population}$}\\

	\# \enspace \textit{Initialise Population} \\

    \For{$i\gets 1, ..., \eta_{p}$}{	    
    {$ Pop.position \gets \text{randomly-generated chromosomes}$}\\
    {$ Pop.cost \gets \text{calculated costs given each chromosome}$}\\
        }\
	{$Pop \gets \text{Sort population ($Pop$);}$}\\
	\# \enspace \textit{Main Loop} \\
	{$\eta_{c} \gets \text{Size of crossover (based on $\theta$)}$}\\
	{$\eta_{m} \gets \text{Size of mutation (based on $\mu$)}$}\\
    \For{$i\gets 1, ..., I^{Max}$}{	    
	\# \enspace \textit{Crossover operation} \\
	    \textit{Calculate probabilities based on $Pr(s\in Pop)= \frac{exp^{(-\beta)* \frac{J_s}{LargestCost}}}{\sum_{k=1}^{\eta_{p}} exp^{(-\beta)* \frac{J_{k}}{LargestCost}}}$} \\

    \For{$i\gets 1, ..., \eta_{c} $}{	    
    \textit{Select two parents (P1 and P2) based on a Roulette Wheel method given probabilities measured above;} \\    	    
    {$ [Offspring(i,1).position,  Offspring(i,2).position]\gets \text{CrossoverFn(P1.position, P2.position)}$}\\
    {$ [Offspring(i,1).cost,  Offspring(i,2).cost]\gets \text{CostFn(P1.position, P2.position)}$}\\
        }\

    	\# \enspace \textit{Mutation operation} \\
   	 \For{$i\gets 1, ..., \eta_{m} $}{	    
 	   \textit{Select one parents (P) based on Roulette Wheel method} \\    	    
  	  {$ [Mutant(i).position]\gets \text{MutationFn(P.position)}$}\\
  	  {$ [Mutant(i).cost]\gets \text{CostFn(P.position)}$}\\
        }\ 	
 	{$Pop \gets \text{ [Pop, Offsprings, and Mutants];}$}\\
  	{$Pop \gets \text{ Sort population and select first $\eta_{p}$ individuals;}$}\\
   	{$BestSolution \gets \text{ Select first chromosome's position, Pop(1).position;}$}\\
   	{$BestCost(i) \gets \text{ Select first chromosome's cost, Pop(1).cost;}$}\\        
        }\	
         {return $\text{Individual (Selected Features);}$}
    \\\hrulefill

	\Fn{$ [O1,O2]=CrossoverFn(P1,P2)$}{
		{$CrossoverMethod \gets \text{ \{Single-Point, Multi-Point, Uniform Crossover\}}$}\\
randomly select one method given probabilities defined for each of them;\\
return two offsprings;\\}\textbf{end}\

	\Fn{$ M=MutationFn(P)$}{
Apply mutation operator;\\
return mutant;\\
}\textbf{end}\

	\Fn{$ J=CostFn(dataset)$}{
Employ ANN;\\
return $\epsilon \times (1+(\Omega \times |X^\prime|))$;\\
}\textbf{end}

    \caption{Pseudo-code for the feature selection model}
    \label{alg:GA}

\end{algorithm*}

\begin{table*}[h]
\caption{Comparing the results of our hybrid model given different parameter setting} 
\centering 
\scriptsize
\begin{tabular}{l c c c rrrrrrrrrrr} 
\hline\hline 
 Crossover Rate & Mutation Rate & Population Size & Neurons &\multicolumn{11}{c}{Corresponding cost for selected solution in different iterations}
\\ [0.5ex]
\hline 
 & &50 
 &10 & $2.0171$ & 2.0171 & 2.0171 & $2.0021$ & $2.0021$ & $...$ & 0.6785 & 0.6785 & 0.6715 & 0.6715 & 0.6642  \\[-0.5ex]
{0.6} & {0.2}&100
& 15 & $1.8264$ & 1.8124 & 1.8124 & $1.8124$ & $1.7023$ & $...$ & 0.5211 & 0.5117 & 0.5011 & 0.5011 & 0.5011  \\[-0.5ex]
 & &150 
 &20 & $1.7111$ & 1.7114 & 1.6617 & $1.6617$ & $1.5668$ & $...$ & 0.5862 & 0.5808 & 0.5808 & 0.5631 & 0.5631   \\[0.5ex]
\hline 
 & &200 
 &10 & $1.8311$ & 1.8311 & 1.8311 & $1.8266$ & $1.8266$ & $...$ & 0.5507 & 0.5507 & 0.5446 & 0.5446 & 0.5446  \\[-0.5ex]
{0.6} & {0.3}&250
& 15 & $1.5182$ & 1.3109 & 1.3109 & $1.1102$ & $1.1102$ & $...$ & 0.4902 & 0.4902 & 0.4902 & 0.4852 & 0.4852  \\[-0.5ex]
 & &300 
 &20 & $1.5330$ & 1.5330 & 1.5330 & $1.4016$ & $1.4001$ & $...$ & 0.5062 & 0.4767 & 0.4767 & 0.4767 & 0.4561   \\[0.5ex]
\hline 
 & &300 
 &10 & $1.0539$ & 1.0539 & 0.9547 & $0.9547$ & $0.9531$ & $...$ & 0.3214 & 0.3112 & 0.3112 & 0.3072 & 0.3072   \\[-0.5ex]
{0.7} & {0.2}&350
& 15 & $0.8413$ & 0.8411 & 0.8231 & $0.8231$ & $0.7877$ & $...$ & 0.2813 & 0.2813 & 0.2813 & 0.2743 & 0.2708   \\[-0.5ex]
 & &400 
 &20 & $0.8613$ & 0.8532 & 0.8152 & $0.7462$ & $0.7462$ & $...$ & 0.2491 & 0.2491 & 0.2491 & 0.2491 & 0.2491   \\[0.5ex]
\hline 
 & &450 
 &10 & $0.6013$ & 0.5182 & 0.5182 & $0.5468$ & $0.5012$ & $...$ & 0.1926 & 0.1926 & 0.1926 & 0.1859 & 0.1859  \\[-0.5ex]
{0.7} & {0.3}&500
& 15 & $0.4165$ & 0.4002 & 0.4002 & $0.3922$ & $0.3774$ & $...$ & 0.1088 & 0.1088 & 0.0991 & 0.0991 & 0.0982  \\[-0.5ex]
 & &550 
 &20 & $0.3911$ & 0.3891 & 0.3891 & $0.3496$ & $0.3347$ & $...$ & 0.1068 & 0.0932 & 0.0932 & 0.0932 & 0.0932   \\[0.5ex]
\hline 
 & &550 
 &10 & $0.2662$ & 0.2662 & 0.2615 & $0.2612$ & $0.2508$ & $...$ & 0.0997 & 0.0997 & 0.0997 & 0.0974 & 0.0974   \\[-0.5ex]
{0.8} & {0.2}&600
& 15 & $0.2227$ & 0.2215 & 0.2016 & $0.2016$ & $0.1835$ & $...$ & 0.0762 & 0.0762 & 0.0742 & 0.0713 & 0.0713  \\[-0.5ex]
 & &650 
 &20 & $0.1844$ & 0.1808 & 0.1808 & $0.1808$ & $0.1808$ & $...$ & 0.0788 & 0.0788 & 0.0781 & 0.0781 & 0.0781   \\[0.5ex]
\hline 
 & &650 
 &10 & $0.1542$ & 0.1523 & 0.1523 & $0.1478$ & $0.1478$ & $...$ & 0.0661 & 0.0661 & 0.0652 & 0.0652 & 0.0652  \\[-0.5ex]
\textbf{0.8} & \textbf{0.3}&\textbf{700}
& \textbf{15} & \textbf{0.1227} & \textbf{0.1214} & \textbf{0.1214} & \textbf{0.1205} & \textbf{0.1205} & $...$ & \textbf{0.0484} & \textbf{0.0484} & \textbf{0.0481} & \textbf{0.0472} & \textbf{0.0415}  \\[-0.5ex]
 & &750 
 &20 & $0.1304$ & 0.1302 & 0.1246 & $0.1246$ & $0.1228$ & $...$ & 0.0493 & 0.0493 & 0.0478 & 0.0478 & 0.0478   \\[0.5ex]
\hline 

\hline 
\end{tabular}
\label{tab:Params}
\end{table*}





\ifCLASSOPTIONcaptionsoff
  \newpage
\fi

\begin{IEEEbiography}[{\includegraphics[width=1in,height=1.25in,clip,keepaspectratio]{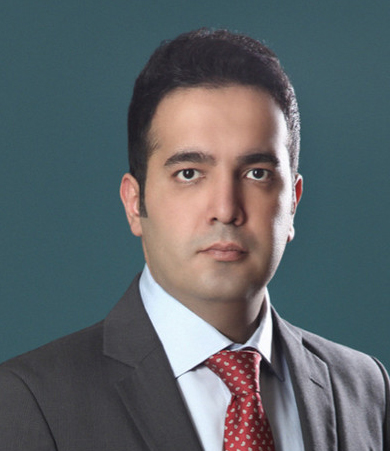}}]{Mohammadhossein Ghahramani}
(S'15-M'18) obtained the B.S. degree and M.S. degree in Information Technology Engineering from Amirkabir University of Technology - Tehran Polytechnic, Iran, and Ph.D. degree in Computer Technology and Application from Macau University of Science and Technology, Macau in 2018. He was a technical manager and senior data analyst of the Information Centre of Institute for Research in Fundamental Sciences from 2008 to 2014. He is currently a Post-Doctoral Research Fellow at University College Dublin (UCD), Ireland. He is also a member of the Insight Centre for Data Analytics at UCD. His research interests include Smart Cities, Machine Learning, Artificial Intelligence, Internet of Things, and big data. Dr. Ghahramani was a recipient of the Best Student Paper Award of 2018 IEEE International Conference on Networking, Sensing and Control. He has served as a reviewer of over ten journals including IEEE Transactions on Cybernetics, IEEE Transactions on Neural Networks and Learning Systems and IEEE Transactions on Industrial Informatics.\end{IEEEbiography}

\begin{IEEEbiography}[{\includegraphics[width=1in,height=1.25in,clip,keepaspectratio]{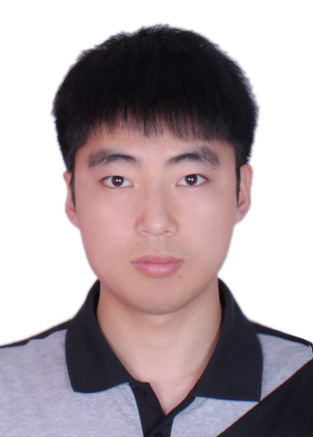}}]{Yan Qiao}
 (M’16) received the B.S. and Ph.D. degrees in Industrial Engineering and Mechanical Engineering from Guangdong University of Technology, Guangzhou, China, in 2009 and 2015, respectively. From Sep. 2014 to Sep. 2015, he was a Visiting Student with the Department of Electrical and Computer Engineering, New Jersey Institute of Technology, Newark, NJ, USA. From Jan. 2016 to Dec. 2017, he was a Post-Doctoral Research Associate with the Institute of Systems Engineering, Macau University of Science and Technology. Since Jan. 2018, he is an Assistant Professor with the Institute of Systems Engineering, Macau University of Science and Technology. He has over 60 publications, including one book chapter and over 30 international journal papers. His research interests include discrete event systems, production planning, Petri nets, scheduling and control. Dr. Qiao was a recipient of the QSI Best Application Paper Award Finalist of 2011 IEEE International Conference on Automation Science and Engineering, the Best Student Paper Award of 2012 IEEE International Conference on Networking, Sensing and Control, and the Best Conference Paper Award Finalist of 2016 IEEE International Conference on Automation Science and Engineering. He has served as a reviewer for a number of journals.
\end{IEEEbiography}

\begin{IEEEbiography}[{\includegraphics[width=1in,height=1.25in,clip,keepaspectratio]{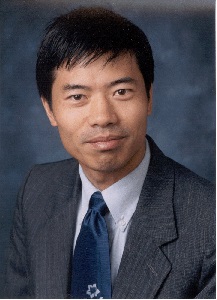}}]{MengChu Zhou}
(S'88-M'90-SM'93-F'03) received his B.S. degree in Control Engineering from Nanjing University of Science and Technology, Nanjing, China in 1983, M.S. degree in Automatic Control from Beijing Institute of Technology, Beijing, China in 1986, and Ph. D. degree in Computer and Systems Engineering from Rensselaer Polytechnic Institute, Troy, NY in 1990.  He joined New Jersey Institute of Technology (NJIT), Newark, NJ in 1990, and is now a Distinguished Professor of Electrical and Computer Engineering. His research interests are in Petri nets, intelligent automation, Internet of Things, big data, web services, and intelligent transportation.  He has over 800 publications including 12 books, over 500 journal papers (over 400 in IEEE transactions), 12 patents and 29 book-chapters. He is the founding Editor of IEEE Press Book Series on Systems Science and Engineering and Editor-in-Chief of IEEE/CAA Journal of Automatica Sinica. He is a recipient of Humboldt Research Award for US Senior Scientists from Alexander von Humboldt Foundation, Franklin V. Taylor Memorial Award and the Norbert Wiener Award from IEEE Systems, Man and Cybernetics Society. He is a life member of the Chinese Association for Science and Technology-USA and served as its President in 1999. He is a Fellow of International Federation of Automatic Control (IFAC), American Association for the Advancement of Science (AAAS) and Chinese Association of Automation (CAA).
\end{IEEEbiography}

\begin{IEEEbiography}[{\includegraphics[width=1in,height=1.25in,clip,keepaspectratio]{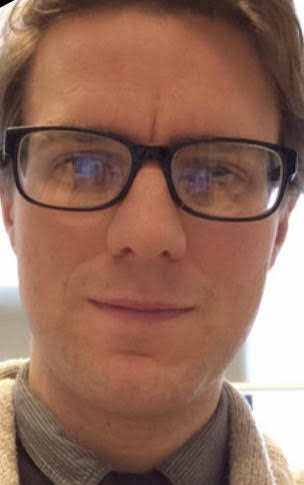}}]{Adrian O Hagan}
is a lecturer and researcher in Statistics and Actuarial Science at University College Dublin (UCD). He holds a degree in Actuarial Science and an MSc and PhD in Statistics from UCD. He uses cutting edge statistical and data analytics techniques to solve real industrial problems in modelling and pricing risk, working with leading insurers and financial institutions. He currently supervises PhD students in Statistical Genetics with Actuarial Applications and Statistics and Actuarial Science, and is currently expanding his research group in the FinTech space. He serves as an examiner for the Institute and faculty of Actuaries and is a referee for several leading Statistics journals. 
\end{IEEEbiography}

\begin{IEEEbiography}[{\includegraphics[width=1in,height=1.25in,clip,keepaspectratio]{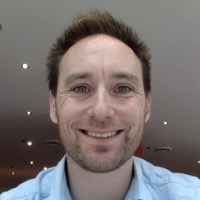}}]{James Sweeney}
is a lecturer and researcher in Statistics at the Royal College of Surgeons  Ireland (RCSI), with a PhD in Statistical Climatology. His research interests range across the fields of spatial analysis, high performance computing and simulation, and statistical applications in medicine and agriculture. Dr. Sweeney's core strengths are in the analysis of extremely large datasets, particularly those comprised of multivariate, spatially indexed data with practical applications including the modelling of house price information in Dublin, as well as evaluating the speed and cost of abrupt climate change. 
\end{IEEEbiography}

\vfill


\begin{thebibliography}{1}

  \bibitem{Jeschke} S. Jeschke, C. Brecher, T. Meisen, D. Ozdemir, and T. Eschert, "Industrial Internet of Things and Cyber Manufacturing Systems," {\em   Industrial Internet of Things, Springer Series in Wireless Technology}, Springer, Cham, pp. 3-19, 2017.

  \bibitem{Lojka} T. Lojka, M. Miskuf M., and I. Zolotova, "Industrial IoT Gateway with Machine Learning for Smart Manufacturing," {\em IFIP Advances in Information and Communication Technology}, pp. 759-766, 2017.

  \bibitem{Ghahramani2020Urban}
M. Ghahramani, M. Zhou and G. Wang, "Urban sensing based on mobile phone data: approaches, applications, and challenges," {\em IEEE/CAA Journal of Automatica Sinica}, vol. 7, no. 3, pp. 627-637, May 2020, doi: 10.1109/JAS.2020.1003120.

  \bibitem{Joze-Tavcar} J. Tavcar and I. Horvath, "A Review of the Principles of Designing Smart Cyber-Physical Systems for Run-Time Adaptation: Learned Lessons and Open Issues," {\em IEEE Transactions on Systems, Man, and Cybernetics: Systems}, vol. 49, pp. 145-158, 2019.

  \bibitem{Bo-Chen} B. Chen, D. W. C. Ho, W. Zhang, and L. Yu, "Distributed Dimensionality Reduction Fusion Estimation for Cyber-Physical Systems Under DoS Attacks," {\em IEEE Transactions on Systems, Man, and Cybernetics: Systems}, vol. 49, pp. 455-468, 2019.

  \bibitem{Peter-Palensky} P. Palensky, E. Widl, and A. Elsheikh, "Simulating Cyber-Physical Energy Systems: Challenges, Tools and Methods," {\em IEEE Transactions on Systems, Man, and Cybernetics: Systems}, vol. 44, pp. 318-326, 2014.

  \bibitem{Yang-Liu} Y. Liu, Y. Peng, B. Wang, S. Yao, and Z. Liu, "Review on cyber-physical systems," {\em IEEE/CAA Journal of Automatica Sinica}, vol. 4, pp. 27-40, 2017.

  \bibitem{Fortino} G. Fortino, F. Messina, D. Rosaci, G. M. Sarne, and C. Savaglio, "A Trust-based Team Formation Framework for Mobile Intelligence in Smart Factories," {\em IEEE Transactions on Industrial Informatics}, 2020, doi: 10.1109/TII.2020.2963910.

  \bibitem{Ghahramani-1} M. Ghahramani, M. C. Zhou, and C. T. Hon, "Analysis of mobile phone data under a cloud computing framework," in {\em Proc. of 14th IEEE International Conference on Networking, Sensing and Control (ICNSC)}, Calabria, Italy, 2017.
  
  \bibitem{Ghahramani-2} M. Ghahramani, M. C. Zhou, and C. T. Hon, "Extracting Significant Mobile Phone Interaction Patterns Based on Community Structures," {\em IEEE Transactions on Intelligent Transportation Systems}, vol. 20, pp. 1031-1041, 2019.
  
  \bibitem{Ghahramani-3} M. Ghahramani, M. C. Zhou, and C. T. Hon, "Mobile Phone Data Analysis: A Spatial Exploration toward Hotspot Detection," {\em IEEE Transactions on Automation Science and Engineering}, vol. 16, pp. 351-362, 2019.
  
  \bibitem{Ghahramani-4} M. H. Ghahramani, M. C. Zhou, and C. T. Hon, "Toward cloud computing QoS architecture: analysis of cloud systems and cloud services," {\em IEEE/CAA Journal of Automatica Sinica}, vol. 4, pp. 6-18, 2017.
  
  \bibitem{D-Bertsekas} D. P. Bertsekas, "Feature-based aggregation and deep reinforcement learning: a survey and some new implementations," {\em IEEE/CAA Journal of Automatica Sinica}, vol. 6, no. 1, pp. 1-31, Jan. 2019.

  \bibitem{Ghahramani2018Spatial} M. Ghahramani, M. Zhou and C. T. Hon, "Spatio-temporal analysis of mobile phone data for interaction recognition," 2018 IEEE 15th International Conference on Networking, Sensing and Control (ICNSC), Zhuhai, 2018, pp. 1-6, doi: 10.1109/ICNSC.2018.8361374.

  \bibitem{Ghahramani2020Intelligent} M. Ghahramani, A. OHagan, M. Zhou and J. Sweeney, "Intelligent Geodemographic Clustering based on Neural Network and Particle Swarm Optimization," {\em IEEE Transactions on Systems, Man and Cybernetics: Systems}, 2020.

  \bibitem{Zhiwei-Gao1} Z. Gao, C. Cecati, and S. X. Ding, "A Survey of Fault Diagnosis and Fault-Tolerant Techniques—Part II: Fault Diagnosis With Knowledge-Based and Hybrid/Active Approaches," {\em IEEE Transactions on Industrial Electronics}, vol. 62, pp. 3768-3774, 2015.

  \bibitem{Alinodehi} S. P. Hoseini Alinodehi, S. Moshfe, M. Saber Zaeimian, A. Khoei, and K. Hadidi, "High-Speed General Purpose Genetic Algorithm Processor," {\em IEEE Transactions on Cybernetics}, vol. 46, pp. 1551-1565, 2016.

  \bibitem{Jiafu-Wan} J. Wan, S. Tang, D. Li, S. Wang, C. Liu, H. Abbas, and A. V. Vasilakos, "A Manufacturing Big Data Solution for Active Preventive Maintenance," {\em IEEE Transactions on Industrial Informatics}, vol. 13, pp. 2039-2047, 2017.

  \bibitem{Semyon-Meerkov} S. M. Meerkov and M. T. Ravichandran, "Combating Curse of Dimensionality in Resilient Monitoring Systems: Conditions for Lossless Decomposition," {\em IEEE Transactions on Cybernetics}, vol. 47, pp. 1263-1272, 2017.

  \bibitem{He-} Q. P. He and J. Wang, "Principal component based k-nearest-neighbor rule for semiconductor process fault detection," in {\em Proc. American Control Conference}, Seattle, USA, 2008, doi: 10.1109/ACC.2008.4586721.

  \bibitem{Cherry-} G.A. Cherry and S.J. Qin, "Principal component based k-nearest-neighbor rule for semiconductor process fault detection," {\em IEEE Transactions on Semiconductor Manufacturing}, vol. 19, pp. 159-172, 2006.

  \bibitem{He-2-} S. He, G. Wang, M. Zhang, and D. Cook, "Multivariate process monitoring and fault identification using multiple decision tree classifiers," {\em International Journal of Production Research}, pp. 3355-3371, 2013.

  \bibitem{He-3-} Q. He and J. Wang, "Fault Detection Using the K-nearest Neighbor Rule for Semiconductor Manufacturing Processes," {\em IEEE Transactions on Semiconductor Manufacturing}, vol. 20, no. 4, pp. 345-354, 2007.

  \bibitem{Verdier-3} G. Verdier and A. Ferreira, "Adaptive Mahalanobis Distance and K-nearest Neighbor Rule for Fault Detection in Semiconductor Manufacturing," {\em IEEE Transactions on Semiconductor Manufacturing}, vol. 24, no. 1, pp. 59-68, 2011.

  \bibitem{Baly-3} R. Baly and H. Hajj, "Wafer Classification Using Support Vector Machines," {\em IEEE Transactions on Semiconductor Manufacturing}, vol. 25, no. 3, pp. 373-383, 2012.

  \bibitem{Kwak-3} J. Kwak, T. Lee, and C. O. Kim, "An Incremental Clustering-based Fault Detection Algorithm for Class-imbalanced Process Data," {\em IEEE Transactions on Semiconductor Manufacturing}, vol. 28, no. 3, pp. 318-328, 2015.

  \bibitem{Zheng-3} Y. Zheng, Q. Liu, E. Chen, Y. Ge, and J. Zhao, "Time series classification using multi-channels deep convolutional neural networks," in {\em Proc. WAIM}, Macau, pp. 298-310, 2014.

  \bibitem{Lee-3} K. Lee, S. Cheon, and C. Kim, "A Convolutional Neural Network for Fault Classification and Diagnosis in Semiconductor Manufacturing Processes," {\em IEEE Transactions on Semiconductor Manufacturing}, vol. 30, pp. 135-142, May 2017.

    \bibitem{Zhang-pca} J. Zhang, H. Chen, S. Chen, and X. Hong, "An Improved Mixture of Probabilistic PCA for Nonlinear Data-Driven Process Monitoring," {\em IEEE Transactions on Cybernetics}, vol. 49, 2019.
  
  
    \bibitem{Xue2016Survey} B. Xue, M. Zhang, W. N. Browne, and Xin Yao, "A Survey on Evolutionary Computation Approaches to Feature Selection," {\em IEEE Transactions on Evolutionary Computation}, vol. 20, pp. 606-626, 2016.   
  
  
    \bibitem{Derrac-first} J. Derrac, S. Garcia, and F. Herrera, "A first study on the use of coevolutionary algorithms for instance and feature selection," in {\em Proc. Hybrid Artificial Intelligence Systems}, Berlin, Germany, 2009.  

    \bibitem{Zamalloa-Feature} M. Zamalloa, G. Bordel, L. J. Rodriguez, and M. Penagarikano, "Feature selection based on genetic algorithms for speaker recognition," in {\em Proc.  IEEE Odyssey Speaker Lang. Recognit. Workshop}, USA, 2006, pp. 1-8. 
  
    \bibitem{Dong-3} L. Dong, S. Chai, B. Zhang, S. K. Nguang, and A. Savvaris, "Stability of a Class of Multiagent Tracking Systems With Unstable Subsystems," {\em IEEE Transactions on Cybernetics}, vol. 47, pp. 2193-2202, 2017.

    \bibitem{Shuo-Wang} S. Wang and X. Yao, "Multiclass Imbalance Problems: Analysis and Potential Solutions," {\em IEEE Transactions on Systems, Man, and Cybernetics, Part B (Cybernetics)}, vol. 42, pp. 1119-1130, 2012.
    
    \bibitem{Liu-Mengchu} H. Liu, M. Zhou and Q. Liu, "An Embedded Feature Selection Method for Imbalanced Data Classification," {\em IEEE/CAA Journal of Automatica Sinica}, 6(3), pp. 703-715, May 2019.  
    
    \bibitem{Q-Kang} Q. Kang, X. Chen, S. Li, and M. C. Zhou, "A Noise-Filtered Under-Sampling Scheme for Imbalanced Classification," {\em IEEE
Transactions on Cybernetics}, 47(12), pp. 4263-4274, Dec. 2018.     
        
  \bibitem{Bunkhumpornpat-1} C. Bunkhumpornpat, K. Sinapiromsaran, and C. Lursinsap, ''DBSMOTE: Density based synthetic minority over-sampling technique,'' {\em Applied Intelligence}, vol. 36, pp. 1-21, 2011.
  
    \bibitem{Xuesong-Zhang} X. Zhang, Y. Zhuang, W. Wang, and W. Pedrycz, "Transfer Boosting with Synthetic Instances for Class Imbalanced Object Recognition," {\em IEEE Transactions on Cybernetics}, vol. 48, 2018.
   
    \bibitem{Gupta-3} A. Gupta, Y. Ong, L. Feng, and K. Tan, "Multiobjective Multifactorial Optimization in Evolutionary Multitasking," {\em IEEE Transactions on Cybernetics}, vol. 49, pp. 1652-1665, 2017.

     \bibitem{Chenping-Hou} C. Hou, F. Nie, X. Li, D. Yi, and Y. Wu, "Joint Embedding Learning and Sparse Regression: A Framework for Unsupervised Feature Selection," {\em IEEE Transactions on Cybernetics}, vol. 44, pp. 793-804, 2013.

     \bibitem{surveyFS} C. Hou, F. Nie, X. Li, D. Yi, and Y. Wu, "A survey on feature selection methods," {\em Computers \& Electrical Engineering}, vol. 40, Issue 1, pp. 16-28, 2014.


\end{thebibliography}
\end{document}